\documentclass[11pt,a4paper]{article}
\usepackage[hyperref]{emnlp-ijcnlp-2019}
\usepackage{times}
\usepackage{latexsym}
\usepackage{graphicx}
\usepackage{mathrsfs}
\usepackage{amsmath}
\usepackage{amssymb}
\usepackage{multirow}
\usepackage[english]{babel}
\usepackage{bm}
\usepackage{url}

\aclfinalcopy 

\title{Event Representation Learning Enhanced with External Commonsense Knowledge}

\author{Xiao Ding, Kuo Liao, Ting Liu\thanks{Corresponding author}, Zhongyang Li, Junwen Duan \\
	Research Center for Social Computing and Information Retrieval \\
	Harbin Institute of Technology, China \\
	{\{xding, kliao, tliu, zyli, jwduan\}@ir.hit.edu.cn}}

\date{}

\begin{document}
\maketitle
\begin{abstract}
  Prior work has proposed effective methods to learn event representations that can capture syntactic and semantic information over text corpus, demonstrating their effectiveness for downstream tasks such as script event prediction. On the other hand, events extracted from raw texts lacks of commonsense knowledge, such as the intents and emotions of the event participants, which are useful for distinguishing event pairs when there are only subtle differences in their surface realizations. To address this issue, this paper proposes to leverage external commonsense knowledge about the intent and sentiment of the event. Experiments on three event-related tasks, i.e., event similarity, script event prediction and stock market prediction, show that our model obtains much better event embeddings for the tasks, achieving 78\% improvements on hard similarity task, yielding more precise inferences on subsequent events under given contexts, and better accuracies in predicting the volatilities of the stock market\footnote{The code and data are available on https://github.com/MagiaSN/CommonsenseERL\_EMNLP\_2019.}.
\end{abstract}

\section{Introduction}

Events are a kind of important \emph{objective} information of the world. Structuralizing and representing such information as machine-readable knowledge are crucial to artificial intelligence \cite{li-etal-2018-generating,li2019story}. The main idea is to learn distributed representations for structured events (i.e. event embeddings) from text, and use them as the basis to induce textual features for downstream applications, such as script event prediction and stock market prediction.

\begin{figure}[!tb]%
	\centering%
	\includegraphics[width=0.48\textwidth]{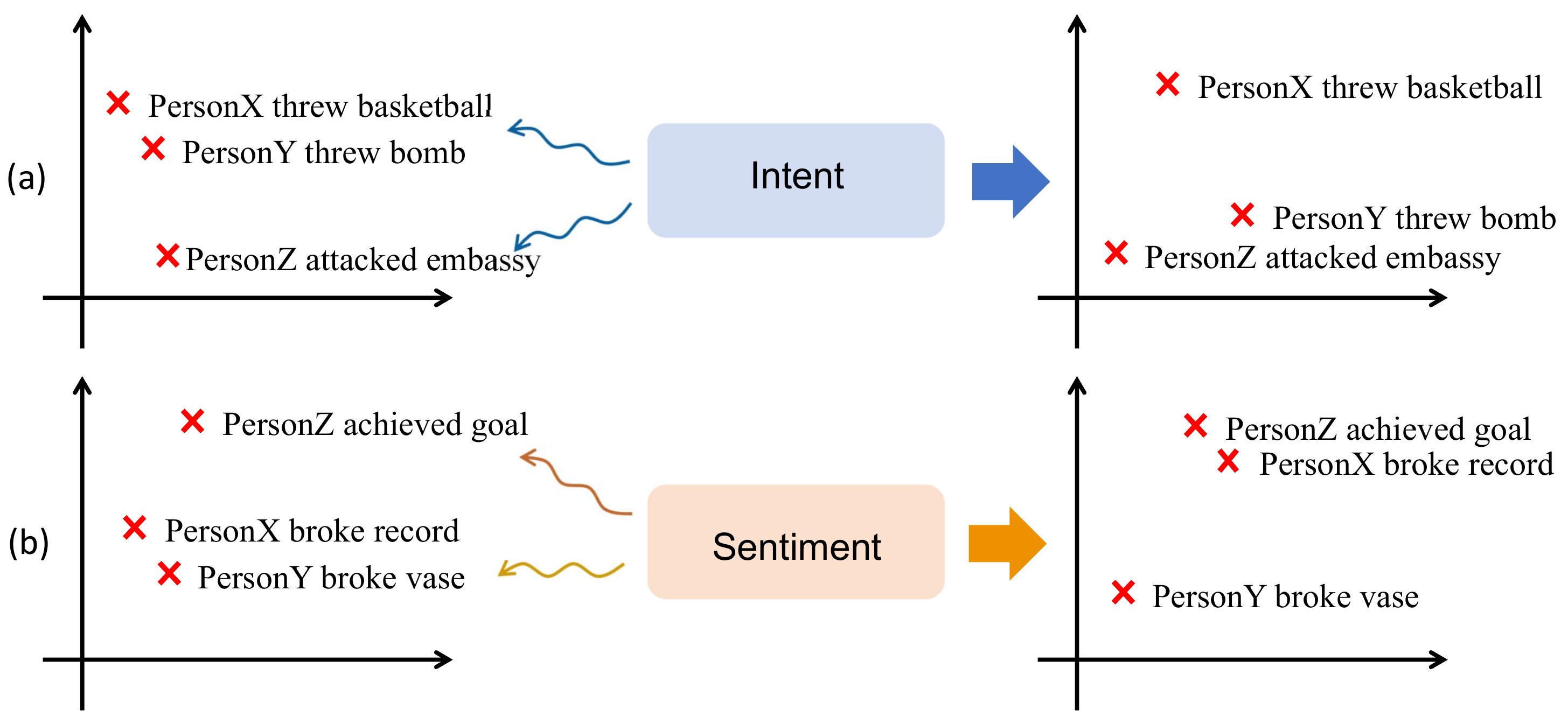}%
	\caption{Intent and sentiment enhanced event embeddings can distinguish distinct events even with high lexical overlap, and find similar events even with low lexical overlap.}%
	\label{fig:exampleintro}%
\end{figure}%

Parameterized additive models are among the most widely used for learning distributed event representations in prior work \cite{granroth2016happens,modi2016event}, which passes the concatenation or addition of event arguments' word embeddings to a parameterized function. The function maps the summed vectors into an event embedding space. Furthermore, \citeauthor{ding2015deep} \shortcite{ding2015deep} and \citeauthor{weber2018event} \shortcite{weber2018event} propose using neural tensor networks to perform semantic composition of event arguments, which can better capture the interactions between event arguments.

\begin{figure*}[htb]
	\centering%
	\includegraphics[width=0.8\textwidth]{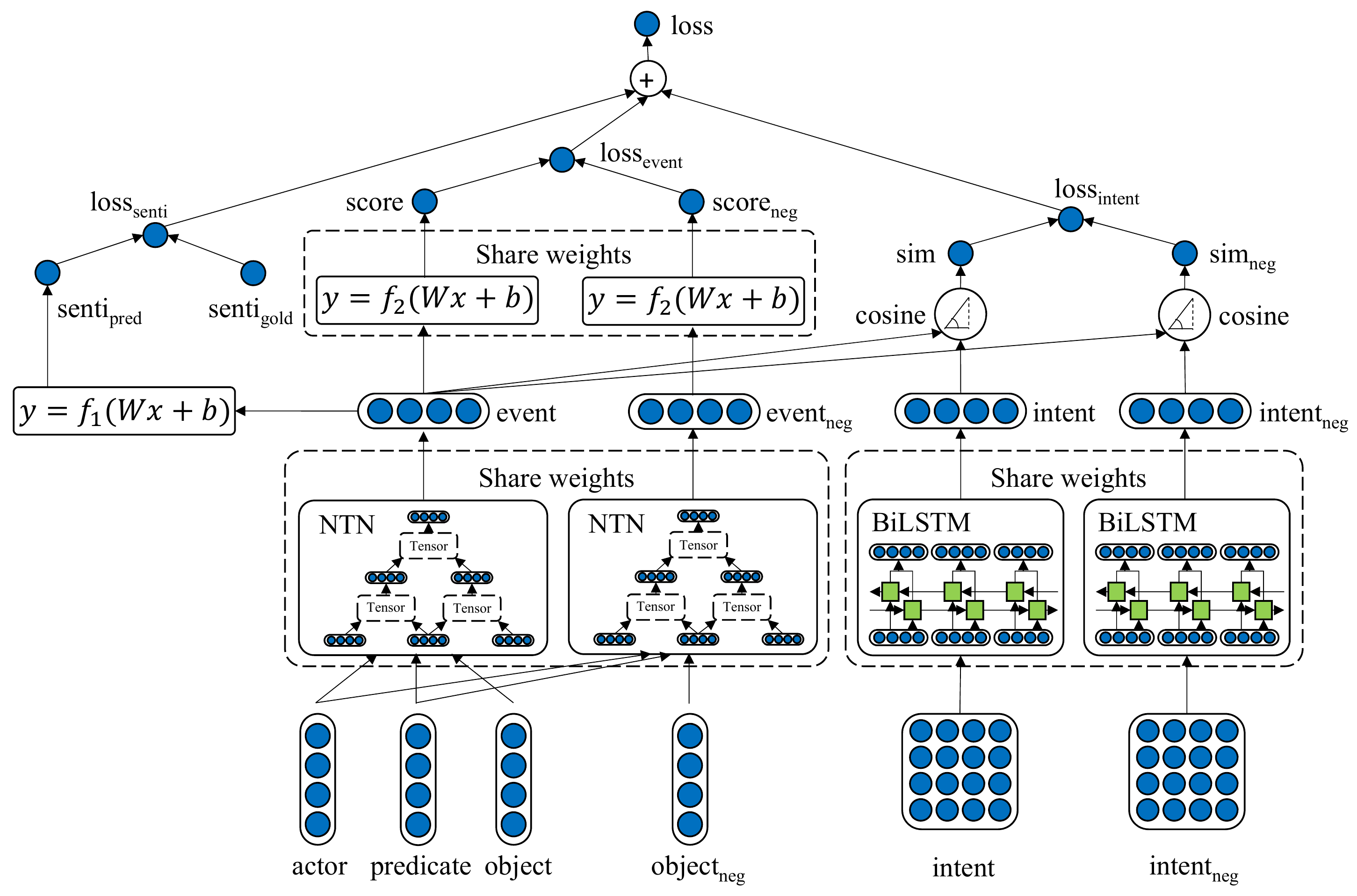}%
	\caption{Architecture of the joint embedding model. $event_{neg}$ refers to the corrupted event tuple, which is derived by replacing each word of the event object with a random word in our dictionary. $intent_{neg}$ is the incorrect intent for the given event, which is randomly selected from the annotated dataset.}%
	\label{fig:framework}%
\end{figure*}%

This line of work only captures shallow event semantics, which is not capable of distinguishing events with subtle differences. On the one hand, the obtained event embeddings cannot capture the relationship between events that are syntactically or semantically similar, if they do not share similar word vectors. For example, as shown in Figure~\ref{fig:exampleintro}~(a), ``\emph{PersonX threw bomb}'' and ``\emph{PersonZ attacked embassy}''. On the other hand, two events with similar word embeddings may have similar embeddings despite that they are quite unrelated, for example, as shown in Figure~\ref{fig:exampleintro}~(b), ``\emph{PersonX broke record}'' and ``\emph{PersonY broke vase}''. Note that in this paper, similar events generally refer to events with strong semantic relationships rather than just the same events.

One important reason for the problem is the lack of the external commonsense knowledge about the mental state of event participants when learning the \emph{objective} event representations. In Figure~\ref{fig:exampleintro}~(a), two event participants ``\emph{PersonY}'' and ``\emph{PersonZ}'' may carry out a terrorist attack, and hence, they have the same intent: ``\emph{to bloodshed}'', which can help representation learning model maps two events into the neighbor vector space. In Figure~\ref{fig:exampleintro}~(b), a change to a single argument leads to a large semantic shift in the event representations, as the change of an argument can result in different emotions of event participants. Who ``\emph{broke the record}'' is likely to be happy, while, who ``\emph{broke a vase}'' may be sad. Hence, intent and sentiment can be used to learn more fine-grained semantic features for event embeddings.

Such commonsense knowledge is not explicitly expressed but can be found in a knowledge base such as Event2Mind \cite{P18-1043} and ATOMIC \cite{sap2018atomic}. Thus, we aim to incorporate the external commonsense knowledge, i.e., \textbf{intent} and \textbf{sentiment}, into the learning process to generate better event representations. Specifically, we propose a simple and effective model to jointly embed events, intents and emotions into the same vector space. A neural tensor network is used to learn baseline event embeddings, and we define a corresponding loss function to incorporate intent and sentiment information.

Extensive experiments show that incorporating external commonsense knowledge brings promising improvements to event embeddings, achieving 78\% and 200\% improvements on hard similarity small and big dataset, respectively. With better embeddings, we can achieve superior performances on script event prediction and stock market prediction compared to state-of-the-art baseline methods.

\section{Commonsense Knowledge Enhanced Event Representations}
The joint embedding framework is shown in Figure~\ref{fig:framework}. We begin by introducing the baseline event embedding learning model, which serves as the basis of the proposed framework. Then, we show how to model intent and sentiment information. Subsequently, we describe the proposed joint model by integrating intent and sentiment into the original objective function to help learn high-quality event representations, and introduce the training details.

\subsection{Low-Rank Tensors for Event Embedding}
The goal of event embedding is to learn low-dimension dense vector representations for event tuples $E=(A, P, O)$, where $P$ is the action or predicate, $A$ is the actor or subject and $O$ is the object on which the action is performed. Event embedding models compound vector representations over its predicate and arguments representations. The challenge is that the composition models should be effective for learning the interactions between the predicate and the argument. Simple additive transformations are incompetent.

\begin{figure}[tb]
	\centering%
	\includegraphics[width=0.4\textwidth]{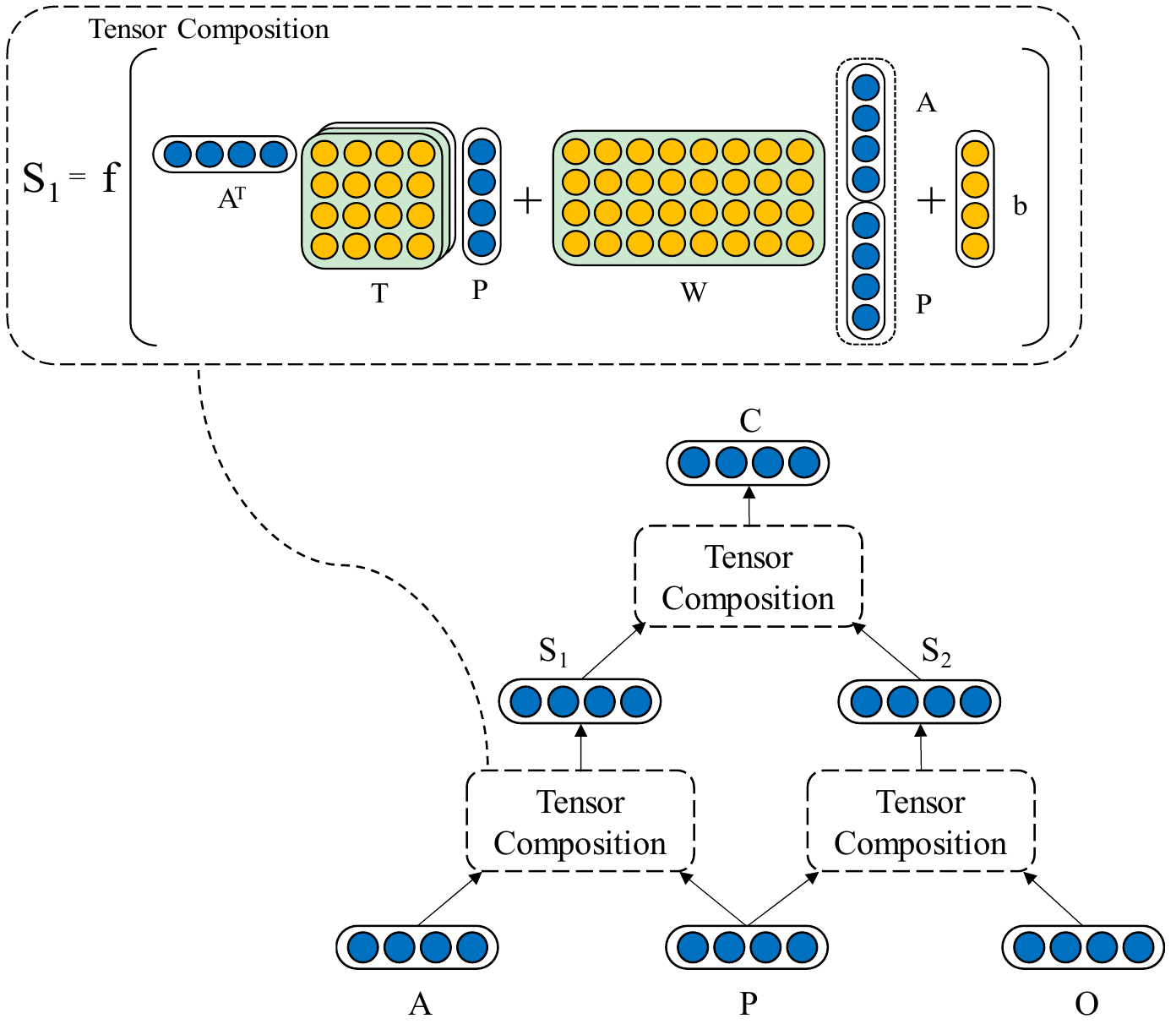}%
	\caption{Baseline event-embedding model.}%
	\label{fig:evb}%
\end{figure}%

We follow \citeauthor{ding2015deep} (\citeyear{ding2015deep}) modelling such informative interactions through tensor composition. The architecture of neural tensor network (NTN) for learning event embeddings is shown in Figure~\ref{fig:evb}, where the bilinear tensors are used to explicitly model the relationship between the actor and the action, and that between the object and the action.

The inputs of NTN are the word embeddings of $A$, $P$ and $O$, and the outputs are event embeddings. We initialized our word representations using publicly available $d$-dimensional ($d=100$) GloVe vectors \cite{pennington2014glove}. As most event arguments consist of several words, we represent the actor, action and object as the average of their word embeddings, respectively.

From Figure~\ref{fig:evb}, $S_1 \in \mathbb{R}^d$ is computed by:

\begin{equation}\small
\begin{split}
\label{equation:tensor}
S_1 & = f\left( A^TT_1^{[1:k]}P + W\left[
\begin{array}{c}
A \\
P \\
\end{array}
\right] + b \right) \\
g(S_1) & = g(A,P)=U^TS_1
\end{split}
\end{equation}

\noindent where $T^{[1:k]}_1 \in \mathbb{R}^{d\times d\times k}$ is a tensor, which is a set of $k$ matrices, each with $d\times d$ dimensions. The bilinear tensor product $A^TT_1^{[1:k]}P$ is a vector $r \in \mathbb{R}^k$, where each entry is computed by one slice of the tensor ($r_i=A^TT_1^{[i]}P, i = 1, \cdots, k$). The other parameters are a standard feed-forward neural network, where $W \in \mathbb{R}^{k \times \it 2d}$ is the weight matrix, $b \in \mathbb{R}^k$ is the bias vector, $U \in \mathbb{R}^k$ is a hyper-parameter and $f=\it tanh$ is a standard nonlinearity applied element-wise. $S_2$ and $C$ in Figure~\ref{fig:evb} are computed in the same way as $S_1$. 

\begin{figure}%
	\centering%
	\includegraphics[width=0.485\textwidth]{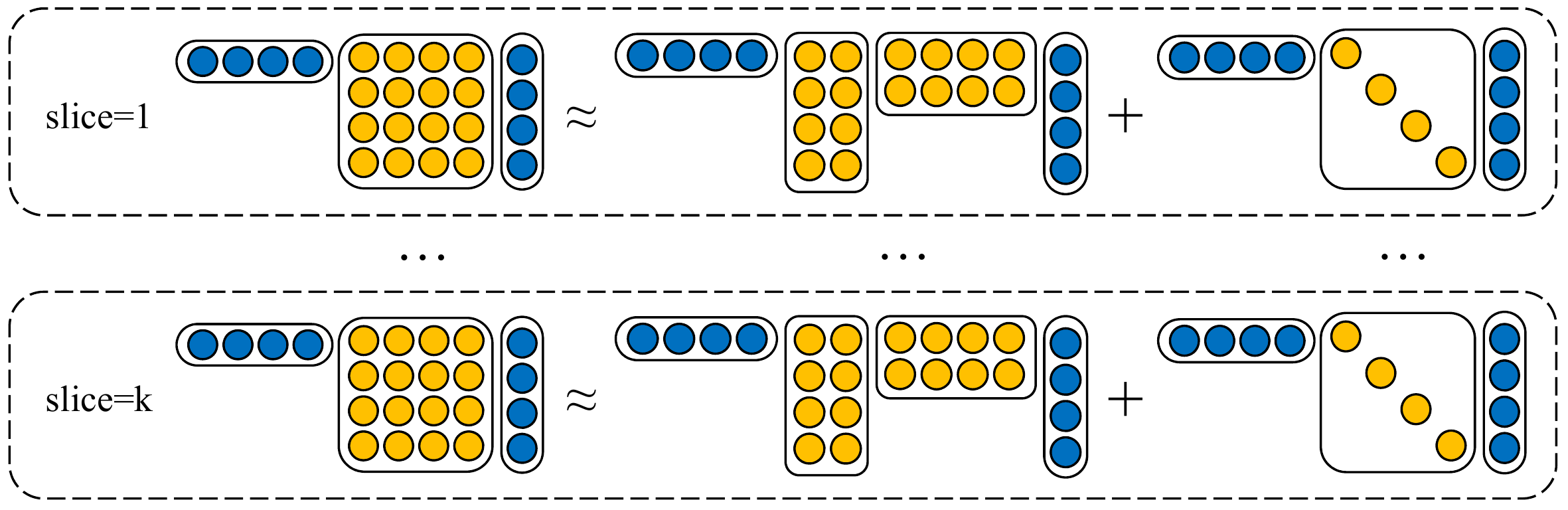}%
	\caption{An illustration of low-rank neural tensor network for learning event embeddings.}%
	\label{fig:lowrank}%
\end{figure}%

One problem with tensors is \emph{curse of dimensionality}, which limits the wide application of tensors in many areas. It is therefore essential to approximate tensors of higher order in a compressed scheme, for example, a low-rank tensor decomposition. To decrease the number of parameters in standard neural tensor network, we make low-rank approximation that represents each matrix by two low-rank matrices plus diagonal, as illustrated in Figure~\ref{fig:lowrank}. Formally, the parameter of the $i$-th slice is $T_{appr}^{[i]}=T^{[i_1]}\times T^{[i_2]}+diag(t^{[i]})$, where $T^{[i_1]}\in\mathbb{R}^{d\times n}$, $T^{[i_2]}\in\mathbb{R}^{n\times d}$, $t^{[i]}\in\mathbb{R}^d$, $n$ is a hyper-parameter, which is used for adjusting the degree of tensor decomposition. The output of neural tensor layer is formalized as follows.

\begin{equation}\small
\label{equation:lowrank}
S_1=f\left( A^T[T_{appr}]_1^{[1:k]}P + W\left[
\begin{array}{c}
A \\
P \\
\end{array}
\right] + b \right),
\end{equation}

\noindent where $[T_{appr}]_1^{[1:k]}$ is the low-rank tensor that defines multiple low-rank bilinear layers. $k$ is the slice number of neural tensor network which is also equal to the output length of $S_1$. 

We assume that event tuples in the training data should be scored higher than corrupted tuples, in which one of the event arguments is replaced with a random argument. Formally, the corrupted event tuple is $E^r=(A^r, P, O)$, which is derived by replacing each word in $A$ with a random word $w^r$ in our dictionary $\mathcal{D}$ (which contains all the words in the training data) to obtain a corrupted counterpart $A^r$. We calculate the \emph{margin loss} of the two event tuples as:

\begin{equation}\small
\label{equation:loss}
\mathcal{L}_\mathcal{E}=loss(E,E^r)=\max(0,1-g(E)+g(E^r))+\lambda \lVert \Phi \rVert^2_2,
\end{equation}

\noindent where $\mathit{\Phi}=(T_1, T_2, T_3, W, b)$ is the set of model parameters. The standard $L_2$	regularization is used, for which the weight $\lambda$ is set as 0.0001. The algorithm goes over the training set for multiple iterations. For each training instance, if the loss $loss(E,E^r)=\max(0,1-g(E)+g(E^r))$ is equal to zero, the online training algorithm continues to process the next event tuple. Otherwise, the parameters are updated to minimize the loss using back-propagation \cite{rumelhart1985learning}.

\subsection{Intent Embedding}
Intent embedding refers to encoding the event participants' intents into event vectors, which is mainly used to explain why the actor performed the action. For example, given two events ``\emph{PersonX threw basketball}'' and ``\emph{PersonX threw bomb}'', there are only subtle differences in their surface realizations, however, the intents are totally different. ``\emph{PersonX threw basketball}'' is just for fun, while ``\emph{PersonX threw bomb}'' could be a terrorist attack. With the intents, we can easily distinguish these superficial similar events.

One challenge for incorporating intents into event embeddings is that we should have a large-scale labeled dataset, which annotated the event and its actor's intents. Recently, \citeauthor{P18-1043} \shortcite{P18-1043} and \citeauthor{sap2018atomic} \shortcite{sap2018atomic} released such valuable commonsense knowledge dataset (ATOMIC), which consists of 25,000 event phrases covering a diverse range of daily-life events and situations. For example, given an event ``\emph{PersonX drinks coffee in the morning}'', the dataset labels \emph{PersonX}'s likely intent is ``\emph{PersonX wants to stay awake}''. 

We notice that the intents labeled in ATOMIC is a sentence. Hence, intent embedding is actually a sentence representation learning task. There are a number of deep neural networks for learning distributed sentence vectors, in which bi-directional LSTMs (BiLSTM) \cite{hochreiter1997long} have been a dominant one, giving state-of-the-art performances in language modelling \cite{peters2018deep} and syntactic parsing  \cite{dozat2016deep}. 

Hence, we use the BiLSTM model to learn intent representations, which is composed of two directed LSTM networks. In the forward direction, LSTM processes the input from the left to right and in the backward direction, it processes the input from right to left. In each LSTM component, it reads the input words recurrently using a single hidden state.

Taking the forward LSTM network as an example, we first define the initial state as $\overrightarrow{\bm{h}}^0$. LSTM reads the input word distributed vectors $\bm{x}_0,\bm{x}_1,\dots,\bm{x}_n$, and recurrently calculates the state transition
$\overrightarrow{\bm{h}}^1,\dots,\overrightarrow{\bm{h}}^{n+1}$ the same as \citeauthor{graves2005framewise} (\citeyear{graves2005framewise}).



The backward LSTM networks shares the same recurrent state transition procedure as the forward LSTM network. It also first defines an initial state $\overleftarrow{\bm{h}}^{n+1}$, and reads the input word representations $\bm{x}_n,\bm{x}_{n-1},\dots,\bm{x}_0$, transforming their values to $\overleftarrow{\bm{h}}^n,\overleftarrow{\bm{h}}^{n-1},\dots,\overleftarrow{\bm{h}}^0$, respectively. 

The BiLSTM model uses the concatenated vector of $\overrightarrow{\bm{h}}^t$ and  $\overleftarrow{\bm{h}}^t$ as the hidden vector for the $t$-th word:

\begin{equation}\small
\label{equation:h}
\bm{h}^t=[\overrightarrow{\bm{h}}^t;\overleftarrow{\bm{h}}^t]
\end{equation}

\noindent A single hidden vector representation $\bm{v}_i$ of the input intent can be obtained by concatenating the last hidden states of the two LSTMs:

\begin{equation}\small
\label{equation:g}
\bm{v}_{i}=[\overrightarrow{\bm{h}}^{n+1};\overleftarrow{\bm{h}}^0]
\end{equation}

In the training process, we calculate the similarity between a given event vector $\bm{v}_e$ and its related intent vector $\bm{v}_i$. For effectively training the model, we devise a ranking type loss function as follows:

\begin{equation}\small
\label{equation:lossintent}
\mathcal{L}_\mathcal{I}=max(0,1-cosine(\bm{v}_e,\bm{v}_i)+cosine(\bm{v}_e,\bm{v}'_i))
\end{equation}

\noindent where $\bm{v}'_i$ is the incorrect intent for $\bm{v}_e$, which is randomly selected from the annotated dataset.

\subsection{Sentiment Embedding}

Sentiment embedding refers to encoding the event participants' emotions into event vectors, which is mainly used to explain how does the actor feel after the event. For example, given two events ``\emph{PersonX broke record}'' and ``\emph{PersonX broke vase}'', there are only subtle differences in their surface realizations, however, the emotions of \emph{PersonX} are totally different. After ``\emph{PersonX broke record}'', \emph{PersonX} may be feel happy, while after ``\emph{PersonX broke vase}'', \emph{PersonX} could be feel sad. With the emotions, we can also effectively distinguish these superficial similar events.

We also use ATOMIC \cite{sap2018atomic} as the event sentiment labeled dataset. In this dataset, the sentiment of the event is labeled as words. For example, the sentiment of ``\emph{PersonX broke vase}'' is labeled as ``(\emph{sad, be regretful, feel sorry, afraid})''. We use SenticNet \cite{cambria2018senticnet} to normalize these emotion words ($W=\{w_1, w_2, \dots, w_n\}$) as the positive (labeled as 1) or the negative (labeled as -1) sentiment. The sentiment polarity of the event $P_e$ is dependent on the polarity of the labeled emotion words $P_W$: $P_e=1$, if $\sum_i P_{w_i}>0$, or $P_e=-1$, if $\sum_i P_{w_i}<0$. We use the softmax binary classifier to learn sentiment enhanced event embeddings. The input of the classifier is event embeddings, and the output is its sentiment polarity (positive or negative). The model is trained in a supervised manner by minimizing the cross entropy error of the sentiment classification, whose loss function is given below.

\begin{equation}\small
\label{equation:losssentiment}
\mathcal{L}_\mathcal{S}=-\sum_{x_e\in C}\sum_{l\in L}p^g_l(x_e)\cdot log(p_l(x_e))
\end{equation}

\noindent where $C$ means all training instances, $L$ is the collection of sentiment categories, $x_e$ means an event vector, $p_l(x_e)$ is the probability of predicting $x_e$ as class $l$, $p^g_l(x_e)$ indicates whether class $l$ is the correct sentiment category, whose value is 1 or -1. 

\subsection{Joint Event, Intent and Sentiment Embedding}

Given a training event corpus with annotated intents and emotions, our model jointly minimizes a linear combination of the loss functions on events, intents and sentiment:

\begin{equation}\small
\label{equation:combination}
\mathcal{L}=\alpha \mathcal{L}_\mathcal{E}+\beta \mathcal{L}_\mathcal{I}+\gamma \mathcal{L}_\mathcal{S}
\end{equation}

\noindent where $\alpha, \beta, \gamma \in [0,1]$ are model parameters to weight the three loss functions.

We use the New York Times Gigaword Corpus (LDC2007T07) for pre-training event embeddings. Event triples are extracted based on the Open Information Extraction technology \cite{schmitz2012open}. We initialize input word embeddings with 100 dimensional pre-trained GloVe vectors \cite{pennington2014glove}, and fine-tune initialized word vectors during our model training. We use Adagrad \cite{duchi2011adaptive}  for optimizing the parameters with initial learning rate 0.001 and batch size 128.

\section{Experiments}

We compare the performance of intent and sentiment powered event embedding model with state-of-the-art baselines 
on three tasks: event similarity, script event prediction and stock prediction.



\subsection{Baselines}

We compare the performance of our approach against a variety of event embedding models developed in recent years. These models can be categorized into three groups:

\begin{itemize}
	\item \textbf{Averaging Baseline (Avg)} This represents each event as the average of the constituent word vectors using pre-trained GloVe embeddings \cite{pennington2014glove}.
	
	\item  \textbf{Compositional Neural Network (Comp. NN)} The distributed event representations are learned by concatenating the subject, predicate, and object embeddings into a two layer neural network \cite{modi2013learning,modi2016event,granroth2016happens}.
	
	\item \textbf{Element-wise Multiplicative Composition (EM Comp.)} This method simply concatenates the element-wise multiplications between the verb and its subject/object. 
	
	\item \textbf{Neural Tensor Network} This line of work use tensors to learn the interactions between the predicate and its subject/object \cite{ding2015deep,weber2018event}. According to the different usage of tensors, we have three baseline methods: \textbf{Role Factor Tensor} \cite{weber2018event} which represents the predicate as a tensor, \textbf{Predicate Tensor} \cite{weber2018event} which uses two tensors learning the interactions between the predicate and its subject, and the predicate and its object, respectively, \textbf{NTN} \cite{ding2015deep}, which we used as the baseline event embedding model in this paper, and \textbf{KGEB} \cite{ding2016knowledge}, which incorporates knowledge graph information in NTN.
\end{itemize}

\begin{table*}[!tb]\small
	\centering
	\begin{tabular}{l|cc|c}
		\hline
		\multirow{2}{*}{Method} & \multicolumn{2}{c|}{Hard Similarity (\emph{Accuracy \%})} &  \multirow{2}{*}{Transitive Sentence Similarity ($\rho$)}\\
		& Small Dataset & Big Dataset & \\
		\hline
		Avg & 5.2 & 13.7 & 0.67 \\
		Comp. NN & 33.0 & 18.9 & 0.63 \\
		EM Comp. & 33.9 & 18.7 & 0.57 \\
		Role Factor Tensor & 43.5 & 20.7 & 0.64 \\
		Predicate Tensor & 41.0 & 25.6 & 0.63 \\
		KGEB & 52.6 & 49.8 & 0.61 \\
		\hline
		NTN & 40.0 & 37.0 & 0.60 \\
		NTN+Int & 65.2 & 58.1 & 0.67 \\
		NTN+Senti & 54.8 & 52.2 & 0.61 \\
		NTN+Int+Senti & \textbf{77.4} & \textbf{62.8} & \textbf{0.74}\\ 
		\hline
	\end{tabular}
	\caption{Experimental results on hard similarity dataset and transitive sentence similarity dataset. The small dataset (230 event pairs) of hard similarity task from \citeauthor{weber2018event} \shortcite{weber2018event}, and the big dataset (2,000 event pairs) is annotated by us. The best results are in bold.}
	\label{tab:experimentalresults}
\end{table*}

\subsection{Event Similarity Evaluation}

\subsubsection{Hard Similarity Task}
We first follow \citeauthor{weber2018event} (\citeyear{weber2018event}) evaluating our proposed approach on the hard similarity task. The goal of this task is that similar events should be close to each other in the same vector space, while dissimilar events should be far away with each other. To this end, \citeauthor{weber2018event} (\citeyear{weber2018event}) created two categories of event pairs: 1) event pairs with similar semantics but have very little lexical overlap (e.g., military launch program / army starts initiative); 2) events pairs with dissimilar semantics but have high lexical overlap (e.g., military launch program / military launch missile).

The labeled dataset contains 230 event pairs (115 pairs each of similar and dissimilar types). \citeauthor{weber2018event} (\citeyear{weber2018event}) asked three different annotators to give the similarity/dissimilarity rankings, and only kept those labels that the annotators agreed upon completely. Following \citeauthor{weber2018event} (\citeyear{weber2018event}), for each baseline method and our approach, we obtain the cosine similarity score of the event pairs, and report the fraction of instances where the similar event pair receives a higher cosine score than the dissimilar event pair (referring to \emph{Accuracy} $\in [0,1]$). To evaluate the robustness of our approach, we extend this dataset to 1,000 event pairs (similar and dissimilar events each account for 50\%), and we will release this dataset to the public.

\subsubsection{Transitive Sentence Similarity}

Except for the hard similarity task, we also evaluate our approach on the transitive sentence similarity dataset \cite{kartsaklis2014study}, which contains 108 pairs of transitive sentences: short phrases containing a single subject, object and verb (e.g., company cut cost). 

For each sentence pair, \citeauthor{kartsaklis2014study} (\citeyear{kartsaklis2014study}) asked an annotator to label a similarity score from 1 to 7. For example, sentence pairs such as (medication, achieve, result) and (drug, produce, effect) are labeled with a high similarity value, while sentence pairs such as (author, write, book) and (delegate, buy, land) are annotated with a low similarity value. Since each sentence pair has several annotations, following \citeauthor{weber2018event} (\citeyear{weber2018event}), we use the average annotator score as the gold score\footnote{To directly compare with baseline methods \cite{weber2018event}, this paper compares with averaged annotator scores, other than comparing with every annotator scores.}. We use the Spearman’s correlation ($\rho \in [-1,1]$) to evaluate the cosine similarity given by each model and the gold score.

\subsubsection{Results}

Experimental results of hard similarity and transitive sentence similarity are shown in Table~\ref{tab:experimentalresults}. We find that:

(1) Simple averaging achieved competitive performance in the task of transitive sentence similarity, while performed very badly in the task of hard similarity. This is mainly because hard similarity dataset is specially created for evaluating the event pairs that should be close to each other but have little lexical overlap and that should be farther apart but have high lexical overlap. Obviously, on such dataset, simply averaging word vectors which is incapable of capturing the semantic interactions between event arguments, cannot achieve a sound performance.

(2) Tensor-based compositional methods (NTN, KGEB, Role Factor Tensor and Predicate Tensor) outperformed parameterized additive models (Comp. NN and EM Comp.), which shows that tensor is capable of learning the semantic composition of event arguments.

(3) Our commonsense knowledge enhanced event representation learning approach outperformed all baseline methods across all datasets (achieving 78\% and 200\% improvements on hard similarity small and big dataset, respectively, compared to previous SOTA method), which indicates that commonsense knowledge is useful for distinguishing distinct events.

\begin{table*}[!tb]\small
	\centering
	\begin{tabular}{c|c|p{0.85cm}|p{0.85cm}|c|c|p{0.85cm}|p{0.85cm}}
		\hline
		Event1 & Event 2 & oScore & mScore & Event1 & Event 2 & oScore & mScore\\
		\hline
		man clears test & he passed exam & -0.08 & 0.40 & man passed car & man passed exam & 0.81 & 0.12\\
		he grind corn & cook chops beans & 0.31 & 0.81 & he grind corn & he grind teeth & 0.89 & 0.36 \\
		he made meal & chef cooked pasta & 0.51 & 0.85 & chef cooked pasta & chef cooked books & 0.89 & 0.45 \\
		farmer load truck & person packs car & 0.58 & 0.83 & farmer load truck & farmer load gun & 0.93 & 0.55 \\
		player scored goal & she carried team & 0.19 & 0.44 & she carried bread & she carried team & 0.59 & 0.09\\ 
		\hline
	\end{tabular}
	\caption{Case study of the cosine similarity score changes with incorporating the intent and sentiment. oScore is the original cosine similarity score without intent and sentiment, and mScore is the modified cosine similarity score with intent and sentiment.}
	\label{tab:casestudy}
\end{table*}

\subsubsection{Case Study}
To further analyse the effects of intents and emotions on the event representation learning, we present case studies in Table~\ref{tab:casestudy}, which directly shows the changes of similarity scores before and after incorporating intent and sentiment. For example, the original similarity score of two events ``\emph{chef cooked pasta}'' and ``\emph{chef cooked books}'' is very high (0.89) as they have high lexical overlap. However, their intents differ greatly. The intent of ``\emph{chef cooked pasta}'' is ``\emph{to hope his customer enjoying the delicious food}'', while the intent of ``\emph{chef cooked books}'' is ``\emph{to falsify their financial statements}''. Enhanced with the intents, the similarity score of the above two events dramatically drops to 0.45. For another example, as the event pair ``\emph{man clears test}'' and ``\emph{he passed exam}'' share the same sentiment polarity, their similarity score is boosted from -0.08 to 0.40.

\begin{table}[!tb]\small
	\centering
	\begin{tabular}{lc}
		\hline
		\textbf{Methods} & \textbf{Acc (\%)} \\
		\hline
		SGNN & 52.45 \\
		SGNN+Int & \textbf{53.93}\\
		SGNN+Senti & 53.57\\
		SGNN+Int+Senti & 53.88\\
		\hline
		SGNN+PairLSTM & 52.71\\
		SGNN+EventComp & 54.15\\
		SGNN+EventComp+PairLSTM & 54.93\\
		SGNN+PairLSTM+Int+Senti & 54.14 \\
		SGNN+EventComp+Int+Senti & 55.08\\
		SGNN+EventComp+PairLSTM+Int+Senti & \textbf{56.03}\\
		\hline
	\end{tabular}
	\caption{Results of script event prediction on the test set. The improvement is significant at $p < 0.05$. Acc is short for Accuracy. }
	\label{tab:script}
\end{table}

\subsection{Script Event Prediction}
Event is a kind of important real-world knowledge. Learning effective event representations can be benefit for numerous applications. Script event prediction \cite{chambers2008narrative} is a challenging event-based commonsense reasoning task, which is defined as giving an existing event context, one needs to choose the most reasonable subsequent event from a candidate list. 

Following \citeauthor{li2018constructing} (\citeyear{li2018constructing}), we evaluate on the standard multiple choice narrative cloze (MCNC) dataset \cite{granroth2016happens}. As SGNN proposed by \citeauthor{li2018constructing} (\citeyear{li2018constructing}) achieved state-of-the-art performances for this task, we use the framework of SGNN, and only replace their input event embeddings with our intent and sentiment-enhanced event embeddings.

\begin{figure}%
	\centering%
	\includegraphics[width=0.47\textwidth]{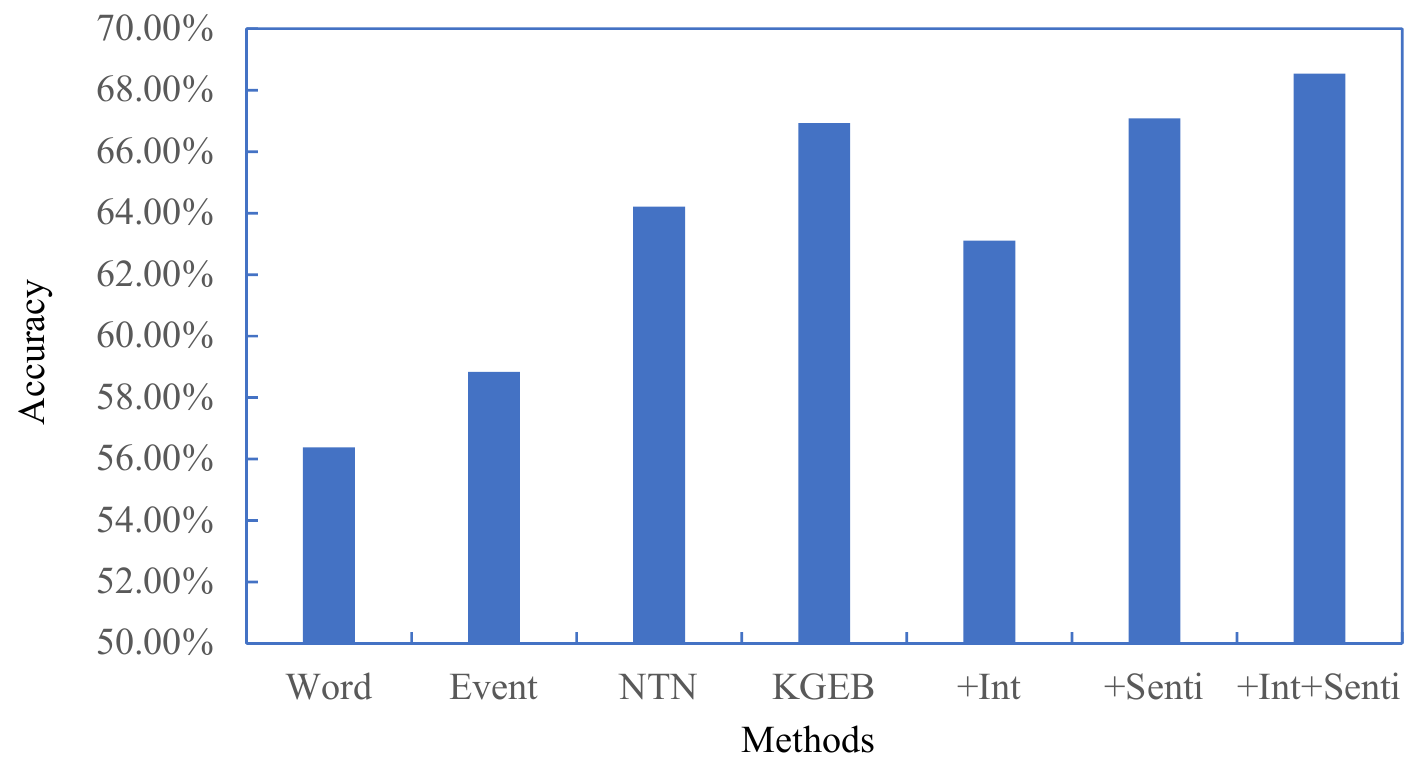}%
	\caption{Experimental results on S\&P 500 index prediction. ``+Int'' means that we encode the intent information into the original event embeddings.}%
	\label{fig:stock}%
\end{figure}%

\citeauthor{D17-1006} (\citeyear{D17-1006}) and \citeauthor{li2018constructing} (\citeyear{li2018constructing}) showed that script event prediction is a challenging problem, and even 1\% of accuracy improvement is very difficult. Experimental results shown in Table~\ref{tab:script} demonstrate that we can achieve more than 1.5\% improvements in single model comparison and more than 1.4\% improvements in multi-model integration comparison, just by replacing the input embeddings, which confirms that better event understanding can lead to better inference results. An interesting result is that the event embeddings only incorporated with intents achieved the best result against other baselines. This confirms that capturing people's intents is helpful to infer their next plan. In addition, we notice that the event embeddings only incorporated with sentiment also achieve better performance than SGNN. This is mainly because the emotional consistency does also contribute to predicate the subsequent event.

\subsection{Stock Market Prediction}
It has been shown that news events influence the trends of stock price movements \cite{luss2012predicting}. As news events affect human decisions and the volatility of stock prices is influenced by human trading, it is reasonable to say that events can influence the stock market.

In this section, we compare with several event-driven stock market prediction baseline methods: (1) \textbf{Word}, \citeauthor{luss2012predicting} \shortcite{luss2012predicting} use bag-of-words represent news events for stock prediction; (2) \textbf{Event}, \citeauthor{ding-EtAl:2014:EMNLP2014} \shortcite{ding-EtAl:2014:EMNLP2014} represent events by subject-predicate-object triples for stock prediction; (3) \textbf{NTN}, \citeauthor{ding2015deep} \shortcite{ding2015deep} learn continues event vectors for stock prediction; (4) \textbf{KGEB}, \citeauthor{ding2016knowledge} \shortcite{ding2016knowledge} incorporate knowledge graph into event vectors for stock prediction. 

Experimental results are shown in Figure~\ref{fig:stock}. We find that knowledge-driven event embedding is a competitive baseline method, which incorporates world knowledge to improve the performances of event embeddings on the stock prediction. Sentiment is often discussed in predicting stock market, as positive or negative news can affect people's trading decision, which in turn influences the movement of stock market. In this study, we empirically show that event emotions are effective for improving the performance of stock prediction (+2.4\%).

\section{Related Work}
Recent advances in computing power and NLP technology enables more accurate models of events with structures. Using open information extraction to obtain structured events representations, we find that the actor and object of events can be better captured \cite{ding-EtAl:2014:EMNLP2014}. For example, a structured representation of the event above can be (Actor = \emph{Microsoft}, Action = \emph{sues}, Object = \emph{Barnes \& Noble}). They report improvements on stock market prediction using their structured representation instead of words as features. 

One disadvantage of structured representations of events is that they lead to increased sparsity, which potentially limits the predictive power. \citeauthor{ding2015deep} \shortcite{ding2015deep} propose to address this issue by representing structured events using \emph{event embeddings}, which are dense vectors. The goal of event representation learning is that similar events should be embedded close to each other in the same vector space, and distinct events should be farther from each other. 

Previous work investigated compositional models for event embeddings. \citeauthor{granroth2016happens} \shortcite{granroth2016happens} concatenate word embeddings of event predicate and event arguments and then learn a nonlinear composition of them into an event representation by a neural network. Event embeddings are further concatenated and fed through another neural network to score how strongly they are expected to appear in the same chain. Modi \shortcite{modi2016event} learn the distributed event representations in a similar way and use that to predict prototypical event orderings. 
\citeauthor{hu2017happens} \shortcite{hu2017happens} learn a distributed event representation for an event description short text rather than a structured event tuple based on a novel hierarchical LSTM model. Then these event embeddings are used to generate a short text describing a possible future subevent. This line of work combines the words in these phrases by passing the concatenation or addition of their word embeddings to a parameterized function that maps the summed vector into event embedding space. However, due to the additive nature of these models, it is difficult to distinguish subtle differences in an event’s surface form. 

To address this issue, \citeauthor{ding2015deep} \shortcite{ding2015deep}, and \citeauthor{weber2018event} \shortcite{weber2018event} propose tensor-based composition models, which can learn the semantic compositionality over the subject, predicate and object of the event by combining them multiplicatively instead of only implicitly or additively, as with standard neural networks.


However, previous work mainly focuses on the nature of the event and lose sight of external commonsense knowledge, such as the intent and sentiment of event participants. This paper proposes to encode intent and sentiment into event embeddings, such that we can obtain a kind of more powerful event representations.

\section{Conclusion}
Understanding events requires effective representations that contain commonsense knowledge. High-quality event representations are valuable for many NLP downstream applications. This paper proposed a simple and effective framework to incorporate commonsense knowledge into the learning process of event embeddings. Experimental results on event similarity, script event prediction and stock prediction showed that commonsense knowledge enhanced event embeddings can improve the quality of event representations and benefit the downstream applications.

\section*{Acknowledgments}

\noindent We thank the anonymous reviewers for their constructive comments, and gratefully acknowledge the support of the National Key Research and Development Program of China (SQ2018AAA010010), the National Key Research and Development Program of China (2018YFB1005103), the National Natural Science Foundation of China (NSFC) via Grant 61702137.


\bibliography{acl2019}
\bibliographystyle{acl_natbib}

\end{document}